\documentclass[11pt,a4paper]{article}
\usepackage[hyperref]{naaclhlt2018}
\usepackage{hyperref}
\usepackage{times}
\usepackage{latexsym}

\usepackage{url}

\usepackage[T1]{fontenc}

\usepackage{graphicx}
\usepackage{eurosym}
\usepackage{tabularx}
\graphicspath{ {images/} }
\usepackage{listings}
\usepackage{stfloats}
\usepackage{hhline}
\usepackage{csquotes}
\usepackage{array}
\usepackage[super]{nth}
\usepackage[inline,shortlabels]{enumitem}
\usepackage{subfig}
\usepackage{amsmath}
\usepackage{multirow}
\usepackage{float}
\usepackage[scientific-notation=true]{siunitx}
\usepackage[colorinlistoftodos,prependcaption,textsize=tiny]{todonotes}
\usepackage{soul}
\usepackage{xargs}                      % Use more than one optional parameter in a new commands
\usepackage{xcolor}
\usepackage{mdframed}

\aclfinalcopy % Uncomment this line for all SemEval submissions

\newcommandx{\improvement}[2][1=]{\todo[linecolor=pink,backgroundcolor=pink!25,bordercolor=pink,#1]{#2}}

\title{NTUA-SLP at SemEval-2018 Task 2: Predicting Emojis using RNNs with Context-aware Attention}
\author{ 
	Christos Baziotis$^{1,3}$,  Nikos Athanasiou$^1$ \\ 
	{\bf Georgios Paraskevopoulos$^{1,4}$, Nikolaos Ellinas$^1$} \\
    {\bf Athanasia Kolovou$^{1,2}$, Alexandros Potamianos$^{1,4}$}
    \\\\
	$^1$School of ECE, National Technical University of Athens, Athens, Greece \\
	$^2$ Department of Informatics, University of Athens, Athens, Greece \\
	$^3$ Department of Informatics, Athens University of Economics and Business, Athens, Greece \\
	$^4$ Behavioral Signal Technologies, Los Angeles, CA\\       
	{\tt cbaziotis@mail.ntua.gr, el12074@central.ntua.gr} \\
	{\tt geopar@central.ntua.gr, nellinas@central.ntua.gr} \\
	{\tt akolovou@di.uoa.gr, potam@central.ntua.gr}
}

\date{2018}

\begin{document}
\maketitle
\begin{abstract}
	In this paper we present a deep-learning model that competed at SemEval-2018 Task 2 \enquote{Multilingual Emoji Prediction}. 
    We participated in subtask A, in which we are called to predict the most likely associated emoji in English tweets. 
	The proposed architecture relies on a Long Short-Term Memory network, augmented with an attention mechanism, that conditions the weight of each word, on a ``context vector'' which is taken as the aggregation of a tweet's meaning.
% 	Furthermore, we apply a sophisticated preprocessing pipeline, customized for Twitter content. 
    Moreover, we initialize the embedding layer of our model, with word2vec word embeddings, pretrained on a dataset of 550 million English tweets.
    Finally, our model does not rely on hand-crafted features or lexicons and is trained end-to-end with back-propagation.
We ranked \nth{2} out of 48 teams.
\end{abstract}

\section{Introduction}
% Social media have transformed written language, introducing new constructs, such as emojis and emoticons\cite{wang2015sentiment,barbieri2016does}.
Emojis play an important role in textual communication, as they function as a substitute for non-verbal cues, that are taken for granted in face-to-face communication, thus allowing users to convey emotions by means other than words.
Despite their large appeal in text, they haven't received much attention until recently. Former works, mostly consider their semantics \cite{aoki2011method,espinosa2016revealing,barbieri2016does,barbieri2016cosmopolitan,ljubevsic2016global,eisner2016emoji2vec} and only recently their role in social media was explored \cite{barbieri2017emojis,cappallo2018new}. 
In SemEval-2018 Task 2: \enquote{Multilingual Emoji Prediction} \cite{semeval2018task2}, given a tweet, we are asked to  predict its most likely associated emoji. 
% Subtask A involves English tweets, while subtask B Spanish tweets. 
% The motivation behind this task is that emoji prediction can enhance natural language understanding in terms of sentiment, emotion \cite{felbo2017using} or irony.

\begin{figure}[!t]
\begin{mdframed}
  \captionsetup{farskip=0pt} % no gap at the top
  \hspace{10pt}\includegraphics[scale=0.55, page=13]{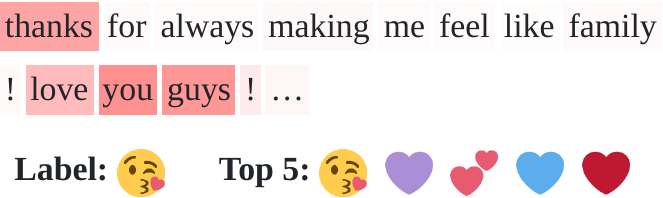}\label{fig:intro_att_1}
  \hfill
  \vskip 14pt
  \hspace{10pt}\includegraphics[scale=0.55, page=39]{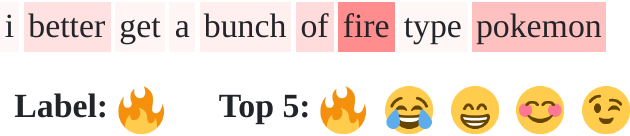}\label{fig:intro_att_2}
\end{mdframed}

\caption{Attention heat-map visualization. 
The color intensity corresponds to the weight given to each word by the self-attention mechanism.}
\label{fig:intro-att}
\end{figure}

In this work, we present a near state of the art approach for predicting emojis in tweets, which outperforms the best present work \cite{barbieri2017emojis}. For this purpose, we employ an LSTM network augmented with a context-aware self-attention mechanism, producing a feature representation used for classification.  Moreover, the attention mechanism helps us make our model's behavior more interpretable, by examining the distribution of the attention weights for a given tweet. To this end, we provide visualizations with the distributions of the attention weights.
\section{Overview} \label{sec:over}
\begin{figure}[t]
	\captionsetup{farskip=0pt} % no gap at the top
	\centerline{\includegraphics[width=1.0\columnwidth]{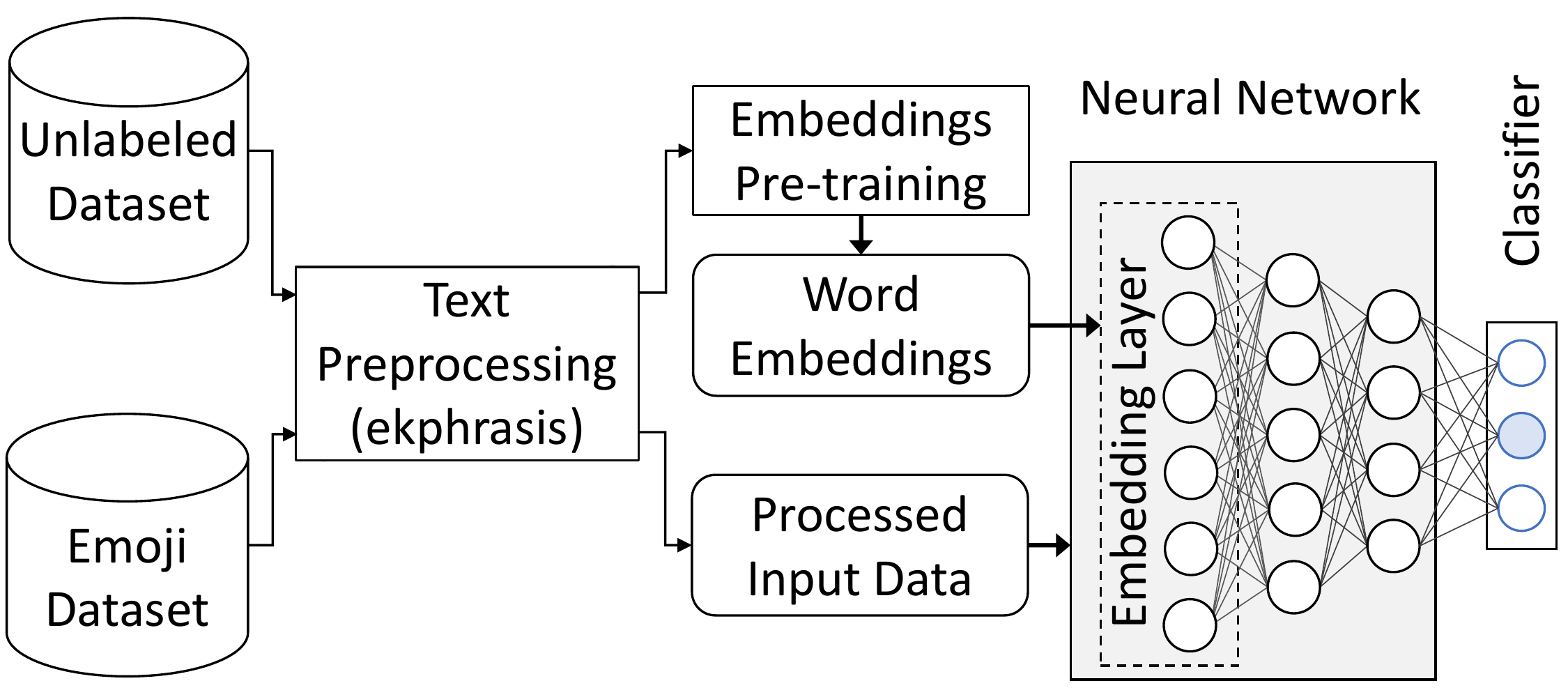}}
	\caption{High-level overview of our approach}
	\label{fig:over}
\end{figure}

Figure~\ref{fig:over} provides a high-level overview of our approach that consists of three main steps: 
\begin{enumerate*}[(1)]
	\item The \emph{text preprocessing step}, which is common both for unlabeled data and the task's dataset,
	\item the \emph{word embeddings pre-training step}, where we train custom word embeddings on a big collection of unlabeled Twitter messages and
	\item the \emph{model training step} where we train the deep learning model.
\end{enumerate*}

\begin{figure}[h]
	\small
    \vspace{-10pt}
	\begin{center}$
		\begin{array}{ccccc}
			\includegraphics[height=0.37cm,width=0.37cm]{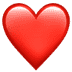} &
			\includegraphics[height=0.37cm,width=0.37cm]{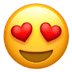} &
			\includegraphics[height=0.37cm,width=0.37cm]{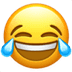} &
			\includegraphics[height=0.37cm,width=0.37cm]{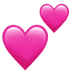} &
			\includegraphics[height=0.37cm,width=0.37cm]{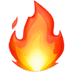}  \\
			22.42\% & 10.34\% & 10.18\% & 5.48\% & 4.91\% \vspace{-4pt}\\\\
            
			\includegraphics[height=0.37cm,width=0.37cm]{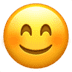} &
            \includegraphics[height=0.37cm,width=0.37cm]{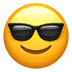} &
			\includegraphics[height=0.37cm,width=0.37cm]{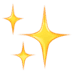} &
			\includegraphics[height=0.37cm,width=0.37cm]{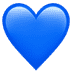} & 
			\includegraphics[height=0.37cm,width=0.37cm]{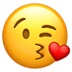} \\ 
			4.67\%  & 4.26\%  & 3.64\%  & 3.40\% & 3.23\% \vspace{-4pt}\\\\
            
			\includegraphics[height=0.37cm,width=0.37cm]{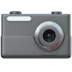} &
			\includegraphics[height=0.37cm,width=0.37cm]{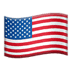} &
			\includegraphics[height=0.37cm,width=0.37cm]{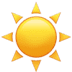} &
			\includegraphics[height=0.37cm,width=0.37cm]{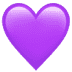} & 
			\includegraphics[height=0.37cm,width=0.37cm]{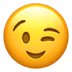} \\
			3.22\%  & 3.04\%  & 2.90\%  & 2.60\% & 2.70\% \vspace{-4pt}\\\\
            
			\includegraphics[height=0.37cm,width=0.37cm]{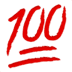} &
			\includegraphics[height=0.37cm,width=0.37cm]{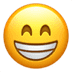} &  
			\includegraphics[height=0.37cm,width=0.37cm]{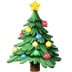} &
			\includegraphics[height=0.37cm,width=0.37cm]{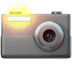} & 
			\includegraphics[height=0.37cm,width=0.37cm]{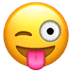}\\
			2.68\%  & 2.61\%  & 2.58\%  & 2.66\% & 2.48\% 
		\end{array}$
	\end{center}
	\caption{Distribution of emoji (class) labels.}
	\label{fig:emoj}
\end{figure}

\noindent\textbf{Task definition}. In subtask A, given an English tweet, we are called to predict the most likely associated emoji, from the 20 most frequent emojis in English tweets according to \cite{barbieri2017emojis}. The training dataset consists of 500k tweets, retrieved from October 2015 to February 2017 and geolocalized in the United States. 
Fig.~\ref{fig:emoj} shows the classes (emojis) and their distribution.
% In the task we are given a Twitter message and we are asked to predict the emoji. The emoji labels are the 20 most frequent in the English language, based on the data collected in \cite{barbieri2017emojis}. An overview of the distributions of the classes is shown in Figure \ref{fig:emoj}:

\subsection{Data}\label{sec:data}
\noindent\textbf{Unlabeled Dataset}. We collected a dataset of 550 million archived English Twitter messages, from Apr. 2014 to Jun. 2017. This dataset is used for (1) calculating word statistics needed in our text preprocessing pipeline (Section \ref{sec:prep}) and (2) training word2vec word embeddings.

\noindent\textbf{Word Embeddings}.
% Word embeddings are dense vector representations of words~\cite{collobert2008, mikolov2013}, capturing their semantic and syntactic information.
We leverage our unlabeled dataset to train Twitter-specific word embeddings. 
We use the \textit{word2vec} ~\cite{mikolov2013} algorithm, with the skip-gram model, negative sampling of 5 and minimum word count of 20, utilizing the Gensim's~\cite{rehurek_lrec} implementation. The resulting vocabulary contains $800,000$ words. 
The pre-trained word embeddings are used for initializing the first layer (embedding layer) of our neural networks.

{\setlength\extrarowheight{0.2em}
	\begin{table*}[!hb]
		\captionsetup{farskip=0pt} % no gap at the topx
		\small
		\begin{tabularx}{\linewidth}{ |c|X| }
			\hline
			original & The *new* season of \#TwinPeaks is coming on May 21, 2017. CANT WAIT \textbackslash o/ !!! \#tvseries \#davidlynch :D \\ 
			\hline
			processed & the new <emphasis> season of <hashtag> twin peaks </hashtag> is coming on <date> . cant <allcaps> wait <allcaps> <happy> ! <repeated> <hashtag> tv series </hashtag> <hashtag> david lynch </hashtag> <laugh>
			\\ 
			\hline
		\end{tabularx}
		\caption{Example of our text processor}
		\label{table:textpp} 
	\end{table*}
}

\subsection{Preprocessing\footnote{Significant portions of the systems submitted to SemEval 2018 in Tasks 1, 2 and 3, by the NTUA-SLP team are shared, specifically the preprocessing and portions of the DNN architecture. Their description is repeated here for completeness.}} \label{sec:prep}
We utilized the \textit{ekphrasis}\footnote{\url{github.com/cbaziotis/ekphrasis}} tool~\cite{baziotis2017datastories} as a tweet preprocessor. The preprocessing steps included in ekphrasis are: Twitter-specific tokenization, spell correction, word normalization, word segmentation (for splitting hashtags) and word annotation.

%\subsection{Tokenizer}
\noindent\textbf{Tokenization}. 
Tokenization is the first fundamental preprocessing step and since it is the basis for the other steps, it immediately affects the quality of the features learned by the network. 
Tokenization on Twitter is challenging, since there is large variation in the vocabulary and the expressions which are used. There are certain expressions which are better kept as one token (e.g. anti-american) and others that should be split into separate tokens. 
Ekphraris recognizes Twitter markup, emoticons, emojis, dates (e.g. 07/11/2011, April 23rd), times (e.g. 4:30pm, 11:00 am), currencies (e.g. \$10, 25mil, 50\euro), acronyms, censored words (e.g. s**t), words with emphasis (e.g. *very*) and more using an extensive list of regular expressions.

\noindent\textbf{Normalization}. 
After tokenization we apply a series of modifications on extracted tokens, such as spell correction, word normalization and segmentation.
Specifically for word normalization we lowercase words, normalize URLs, emails, numbers, dates, times and user handles (@user). This helps reducing the vocabulary size without losing information.
For spell correction \cite{jurafsky2000} and word segmentation \cite{segaran2009a} we use the Viterbi algorithm. The prior probabilities are initialized using uni/bi-gram word statistics from the unlabeled dataset. 
Table~\ref{table:textpp} shows an example text snippet and the resulting preprocessed tokens.

\subsection{Recurrent Neural Networks}
We model the Twitter messages using Recurrent Neural Networks (RNN). RNNs process their inputs sequentially, performing the same operation, $ h_t=f_W(x_t, h_{t-1}) $, on every element in a sequence,
%\begin{equation} \label{eq:class_weights}
%h_t=f_W(x_t, h_{t-1})
%\end{equation}
where $h_t$ is the hidden state $t$ the time step, and $W$ the network weights. We can see that hidden state at each time step depends on previous hidden states, thus the order of elements (words) is important. This process also enables RNNs to handle inputs of variable length. 

RNNs are difficult to train \cite{pascanu2013a}, because gradients may grow or decay exponentially over long sequences \cite{bengio1994,hochreiter2001}. 
A way to overcome these problems is to use more sophisticated variants of regular RNNs, like Long Short-Term Memory (LSTM) networks~\cite{hochreiter1997} or Gated Recurrent Units (GRU)~\cite{cho2014a}, which ensure better gradient flow through the network. 
% For our models, we use LSTMs.
% Furthermore, we apply gradient clipping \cite{pascanu2013a} as a measure against exploding gradients.

\begin{figure}[!b]
	\captionsetup{farskip=0pt} % no gap at the top
	\centering
	\subfloat[Regular RNN ]{\includegraphics[width=0.48\columnwidth,page=1]{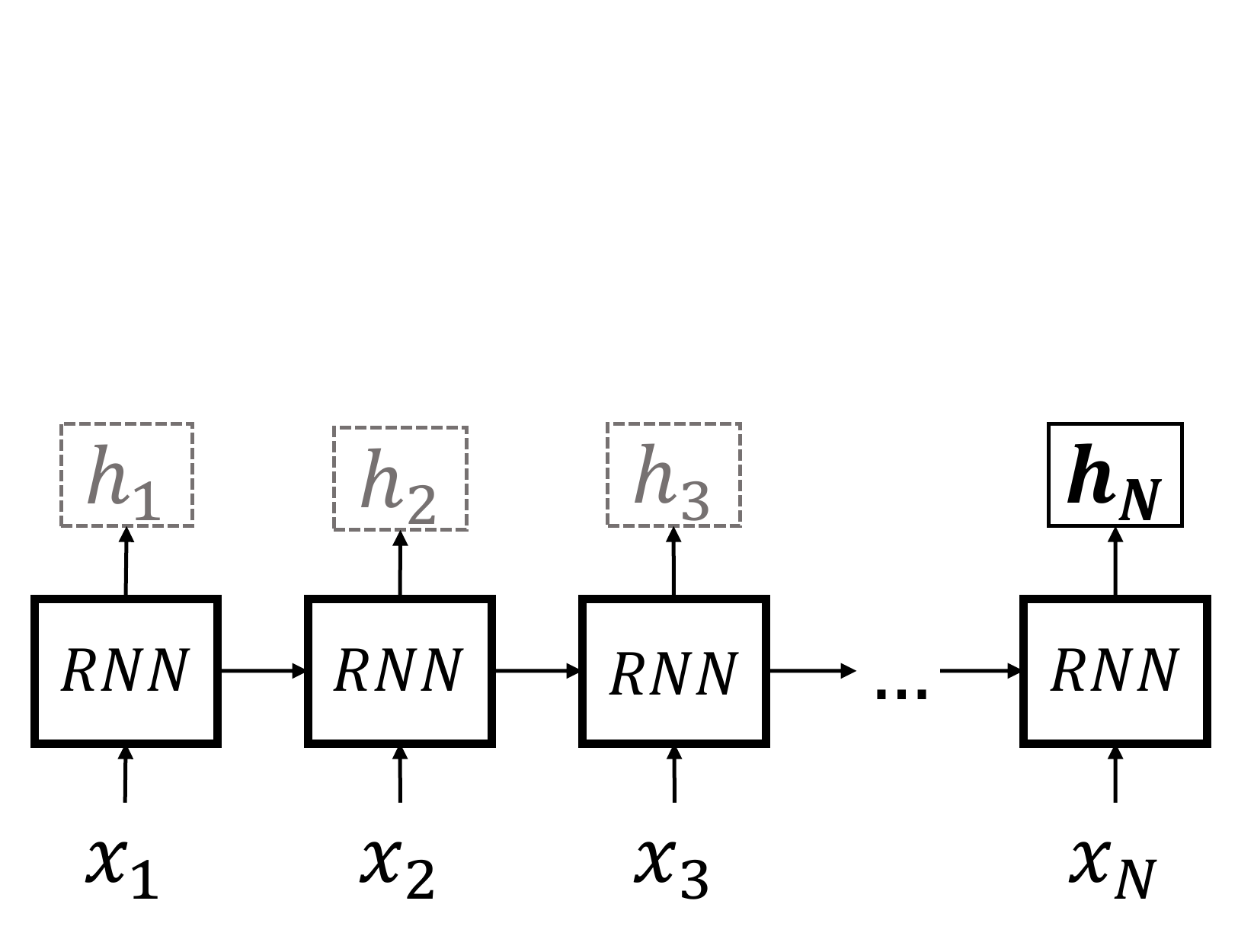}\label{fig:rnn}}
	\hfill
	\subfloat[Attention RNN]{\includegraphics[width=0.48\columnwidth,page=2]{rnn}\label{fig:rnn_att}}
	\caption{Comparison between the regular RNN and the RNN with attention.}
	\label{fig:attention}
\end{figure}

% \subsection{Self-Attention Mechanism}\label{sec:self-att}
\noindent\textbf{Self-Attention Mechanism}.
RNNs update their hidden state $h_i$ as they process a sequence and the final hidden state holds a summary of the information in the sequence. 
In order to amplify the contribution of important words in the final representation, a self-attention mechanism \cite{DBLP:journals/corr/BahdanauCB14} can be used (Fig.~\ref{fig:attention}). 
In normal RNNs, we use as representation $r$ of the input sequence its final state $h_N$. However, using an attention mechanism, we compute $r$ as the convex combination of all $h_i$, with weights $a_i$, which signify the importance of each hidden state. Formally:
$
r = \sum_{i=1}^{N} a_i h_i,  \mbox{where} \sum_{i=1}^{N} a_i = 1, \; \mbox{and} \; a_i > 0.
$

\section{Model Description} \label{models-description}
We use a word-level BiLSTM architecture to model semantic information in tweets. We also propose an attention mechanism, which conditions the weight of $h_i$ on a ``context vector'' that is taken as the aggregation of the tweet meaning.

% \begin{figure*}[!ht]
% 	\captionsetup{farskip=0pt} % no gap at the top
% 	\centering
% 	\includegraphics[width=0.80\textwidth, page=1]{models}
% 	\caption{The proposed model, composed of a BiLSTM with a context-aware self-attention mechanism.}
% 	\label{fig:nn1}
% \end{figure*}

\begin{figure}[!ht]
	\captionsetup{farskip=0pt} % no gap at the top
	\centering
	\includegraphics[width=0.9\columnwidth]{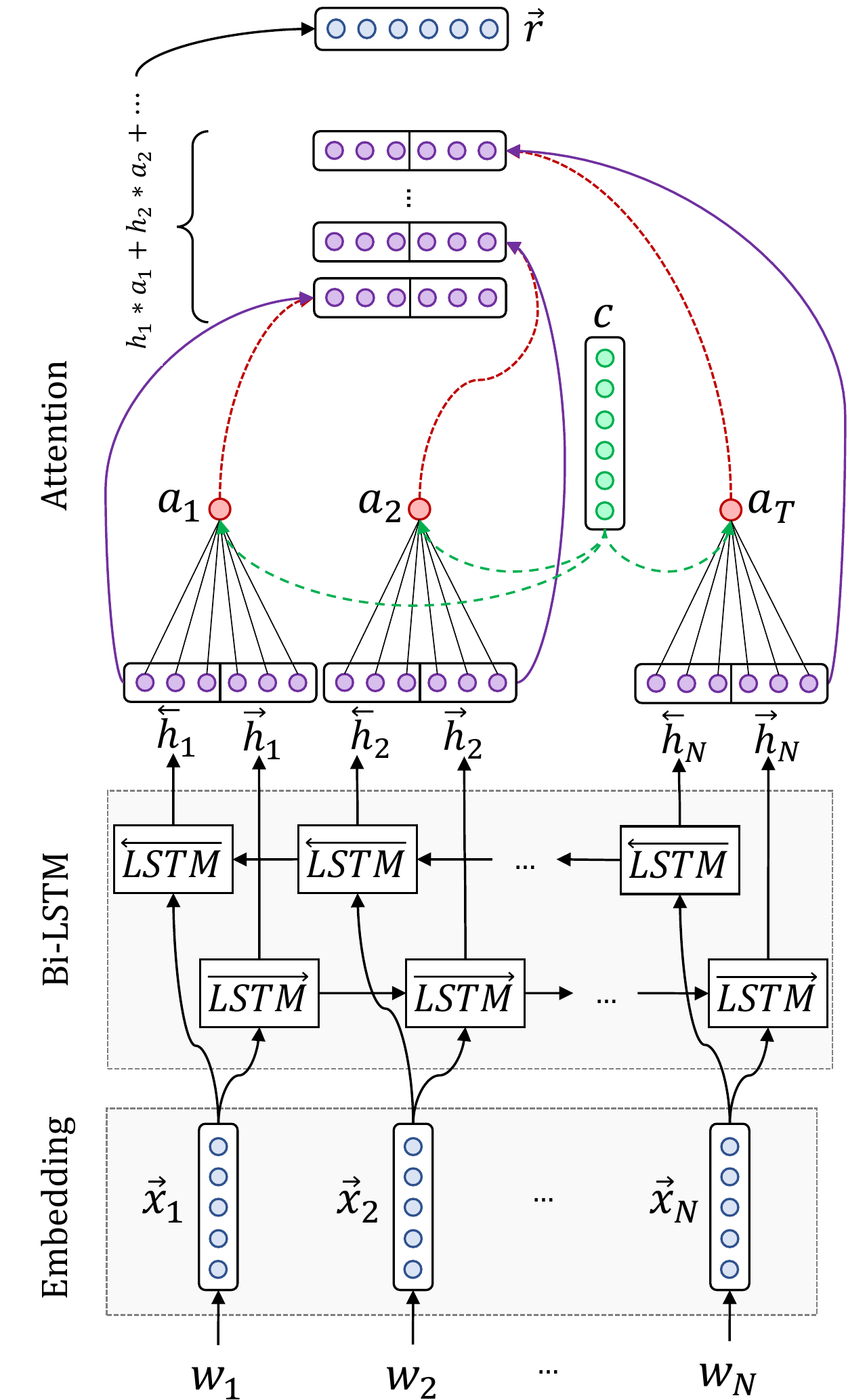}
	\caption{Architecture of the proposed model.}
	\label{fig:nn1}
\end{figure}

\noindent\textbf{Embedding Layer}. The input to the network is a Twitter message, treated as a sequence of words. We use an embedding layer to project the words $w_1,w_2,...,w_N$ to a low-dimensional vector space $ R^W$, where $W$ the size of the embedding layer and $N$ the number of words in a tweet. We initialize the weights of the embedding layer with our pretrained word embeddings.

\noindent\textbf{BiLSTM Layer}. A LSTM takes as input the words of a tweet and produces the word annotations $h_1,h_2,...,h_N$, where $h_i $ is the hidden state of the LSTM at time-step $i$, summarizing  all the information of the sentence up to $w_i$. 
We use bidirectional LSTM (BiLSTM) in order to get word annotations that summarize the information from both directions. A BiLSTM consists of a forward LSTM $ \overrightarrow{f} $  that reads the sentence from $w_1$ to $w_N$ and a backward LSTM $ \overleftarrow{f} $ that reads the sentence from $w_N$ to $w_1$. We obtain the final annotation for each word, by concatenating the annotations from both directions,
$
h_i = \overrightarrow{h_i} \parallel \overleftarrow{h_i}, \quad h_i \in R^{2L}
$
where $ \parallel $ denotes the concatenation operation and $L$ the size of each LSTM.

\noindent\textbf{Context-aware Self-Attention Layer}.
% In order to amplify the contribution of the most informative words, we add a self-attention mechanism on top of the BiLSTM. 
Even though the hidden state $h_i$ of the LSTM captures the local context up to word $i$, in order to better estimate the importance of each word given the context of the tweet we condition hidden state on a context vector. The context vector is taken as the average of $h_i$:  $ c = \dfrac{1}{N}\sum_{1}^{N} h_i $.
The context-aware annotations $u_i$ are obtained as the concatenation of $c$ and $h_i$: $ u_i = h_i\parallel c $.
The attention weights $a_i$ are computed as the softmax of the attention layer outputs $e_i$. $W$ and $b$ are the trainable weights and biases of the attention layer:
\begin{align}
e_i &= tanh(W u_i + b)\label{eq:att_ei}\\
a_i &= \dfrac{exp(e_i)}{\sum_{t=1}^{N} exp(e_t)}\label{eq:att_ai}
\end{align}
The final representation $r$ is again taken as the convex combination of the hidden states.
\begin{equation}
r = \sum_{i=1}^{N} a_i h_i \label{eq:att_r}, \quad r \in R^{2L}
\end{equation} 

% \begin{align}
% u_i &= h_i\parallel c\\
% e_i &= tanh(W u_i + b)\label{eq:att_ei}\\
% a_i &= \dfrac{exp(e_i)}{\sum_{t=1}^{N} exp(e_t)}\label{eq:att_ai}\\
% r &= \sum_{i=1}^{N} a_i h_i \label{eq:att_r}, \quad r \in R^{2L}
% \end{align}
%where $ W $ and $ b $ are the attention layer's weights and bias respectively, optimized during training to assign bigger weights to the most important words of in a sentence.

\noindent\textbf{Output Layer}.
We use the representation $r$ as feature vector for classification and we feed it to a fully-connected softmax layer with $L$ neurons, which outputs a probability distribution over all classes $p_c$ as described in Eq.~\ref{e:outlay}:

\begin{equation}
\label{e:outlay}
p_c = \frac{e^{Wr + b}}{\sum_{i \in [1,L]}(e^{W_i r + b_i})}
\end{equation}
where $W$ and $b$ are the layer's weights and biases.

\subsection{Regularization}\label{sec:reg}
In both models we add Gaussian noise to the embedding layer, which can be interpreted as a random data augmentation technique, that makes models more robust to overfitting.
In addition to that we use dropout \cite{srivastava2014} and we stop training after the validation loss has stopped decreasing (early-stopping).

\section{Experiments and Results} \label{sec:experiments}

\subsection{Experimental Setup}\label{sec:exp_setup}

\noindent\textbf{Class Weights}\label{sec:class_weights}. 
In order to deal with class imbalances, we apply
class weights to the loss function of our models, penalizing more the misclassification of underrepresented
classes. We weight each class by its inverse frequency in the training set.

\noindent\textbf{Training}\label{sec:train}
We use Adam algorithm~\cite{kingma2014} for optimizing our networks, with mini-batches of size 32 and we clip the norm of the gradients~\cite{pascanu2013a} at 1, as an extra safety measure against exploding gradients. For developing our models we used PyTorch \cite{paszke2017automatic} and Scikit-learn \cite{pedregosa2011}.

\noindent\textbf{Hyper-parameters}.
In order to find good hyper-parameter values in a relative short time (compared to grid or random search), we adopt the Bayesian optimization \cite{bergstra2013} approach, performing a time-efficient search in the space of all hyper-parameter values.
The size of the embedding layer is 300, and the LSTM layers 300 (600 for BiLSTM). 
We add Gaussian noise with $\sigma=0.05$ and dropout of 0.1 at the embedding layer and dropout of 0.3 at the LSTM layer.

\noindent\textbf{Results}. The dataset for Task 2 was introduced in \cite{Saggion2017AreEP}, where the authors propose a character level model with pretrained word vectors that achieves an F1 score of $34\%$.  
Our ranking as shown in Table~\ref{tab:res_official} was 2/49, with an F1 score of $35.361\%$, which was the official evaluation metric, while team \texttt{TueOslo} achieved the first position with an F1 score of $35.991\%$. It should be noted that only the first $2$ teams managed to surpass the baseline model presented in \cite{Saggion2017AreEP}.
\begin{table}[b]
\captionsetup{farskip=0pt} % no gap at the topx
\small
\centering
\begin{tabular}{|p{0.2em}|p{1.9cm}|p{0.75cm}|p{0.75cm}|p{0.75cm}|p{0.75cm}|}
\hline 
\multicolumn{1}{|l|}{\textbf{\#}} & \textbf{Team Name} & \multicolumn{1}{c|}{\textbf{Acc}} & \multicolumn{1}{c|}{\textbf{Prec}} & \multicolumn{1}{c|}{\textbf{Rec}} & \multicolumn{1}{c|}{\textbf{F1}} \\ \hline
1 & TueOslo & 47.094 & 36.551 & 36.222 & 35.991 \\ \hline
2 & \textbf{NTUA-SLP} & 44.744 & 34.534 & 37.996 & 35.361 \\ \hline
3 & \textit{Unknown} & 45.548 & 34.997 & 33.572 & 34.018 \\ \hline
4 & Liu Man & 47.464 & 39.426 & 33.695 & 33.665 \\ \hline
\end{tabular}
\caption{Official Results for Subtask A}
\label{tab:res_official}
\end{table}

\begin{table}[tb]
\centering
\small
\begin{tabular}{|l|r|r|r|r|}
\hline
                 & \textbf{f1}     & \textbf{accuracy} & \textbf{recall} & \textbf{precision} \\ \hline
\textbf{BOW}     & 0.3370          & 0.4468            & 0.3321          & 0.3525             \\ \hline
\textbf{N-BOW}   & 0.2904          & 0.4120            & 0.2849          & 0.3150             \\ \hline
\textbf{CA-LSTM} & \textbf{0.3564} & \textbf{0.4482}   & \textbf{0.3885} & \textbf{0.3531}    \\ \hline
\end{tabular}
\caption{Comparison against baselines}
\label{table:comp}
\end{table}

In Table~\ref{table:comp} we compare the proposed Context-Attention LSTM (CA-LSTM) model against 2 baselines: (1) a Bag-of-Words (BOW) model with TF-IDF weighting and (2) a Neural Bag-of-Words (N-BOW) model, where we retrieve the word2vec representations of the words in a tweet and compute the tweet representation as the centroid of the constituent word2vec representations.
Both BOW and N-BOW features are then fed to a linear SVM classifier, with tuned $C=0.6$. The CA-LSTM results in Table~\ref{table:comp} are computed by averaging the results of $10$ runs to account for model variability. Table~\ref{table:comp} shows that BOW model outperforms N-BOW by a large margin, which may indicate that there exist words, which are very correlated with specific classes and their occurrence can determine the classification result. Finally, we observe that CA-LSTM significantly outperforms both baselines.

\begin{figure}[h]
	\captionsetup{farskip=0pt} % no gap at the top
	\centerline{\includegraphics[width=1\columnwidth]{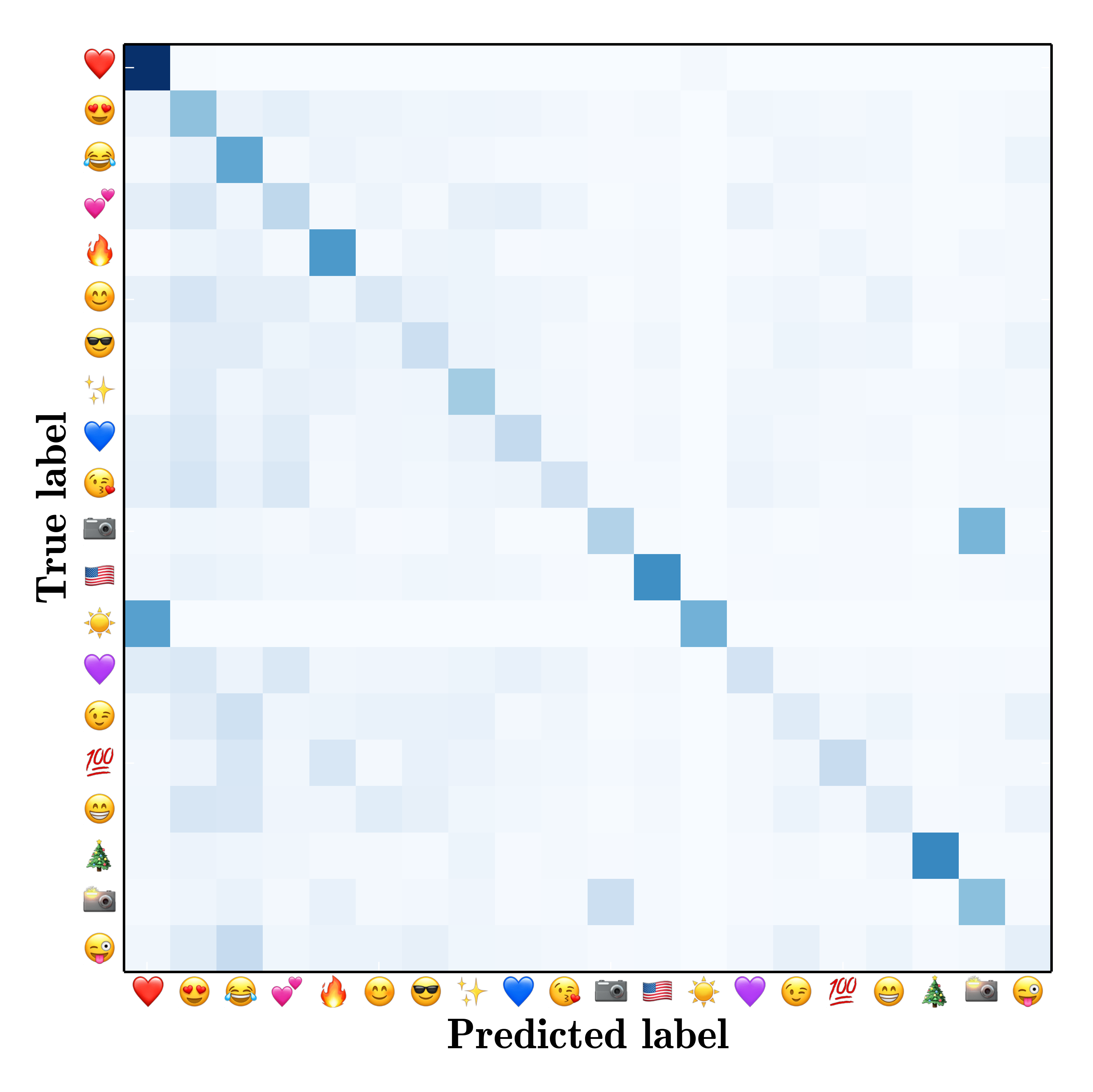}}
	\caption{Confusion matrix}
	\label{fig:conf_matrix}
\end{figure}
Fig. \ref{fig:conf_matrix} shows the confusion matrix for the $20$ emojis. Observe that our model is more likely to misclassify a rare class as an instance of one of the $4$ more frequent classes, 
% which is to be expected 
even after the inclusion of class weights in the loss function (Section \ref{sec:exp_setup}). 
% The imbalances of the dataset are evident in Table \ref{table:emoji_detailed}, where we can see that the most frequent class has $10$ times more occurrences than the most rare, while the $5$ most frequent classes account for the $50\%$ of the total occurrences.
% Note though, that occurrence frequency does not seem to imply higher precision/recall score. 
Furthermore, we observe that heart or face emojis, which are more ambiguous, are easily confusable with each other. However, as expected this in not the case for emojis like the US flag or the Christmas tree, as they are tied with specific expressions.

\begin{figure}[!t]
\begin{mdframed}
  \captionsetup{farskip=0pt} % no gap at the top
  \hspace{5pt}\includegraphics[scale=0.59, page=38]{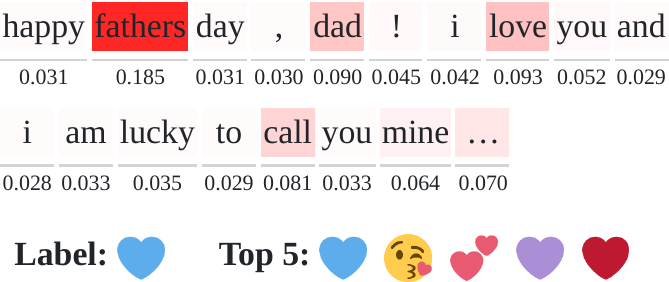}\label{fig:att1}
  \hfill
  \vskip 15pt
%   \hspace{5pt}\includegraphics[scale=0.55, page=28]{heatmaps}\label{fig:att1}
%   \hfill
%   \vskip 15pt
  \hspace{5pt}\includegraphics[scale=0.59, page=34]{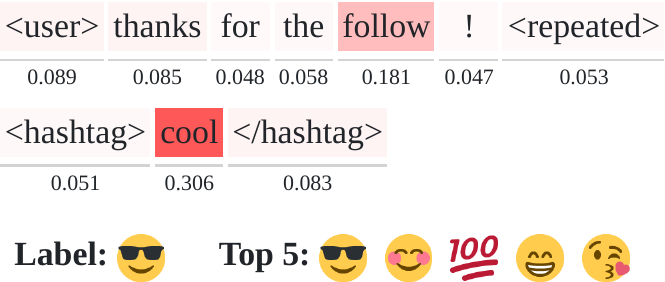}\label{fig:att2}
  \hfill
  \vskip 15pt
  \hspace{5pt}\includegraphics[scale=0.59, page=28]{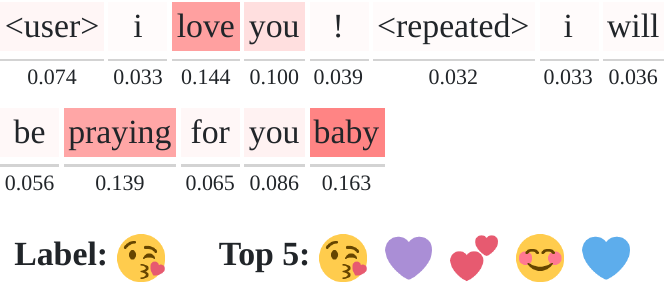}\label{fig:att3}
  \hfill
  \vskip 15pt
  \hspace{5pt}\includegraphics[scale=0.59, page=18]{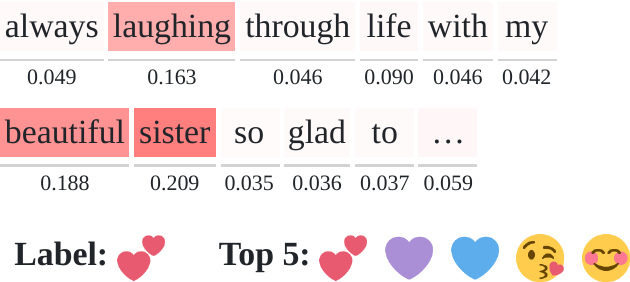}\label{fig:att4}
\end{mdframed}

\caption{Attention Visualizations}\label{fig:attentions}
\end{figure}

\noindent\textbf{Attention Visualization}. 
The attention mechanism not only improves the performance of the model, but also makes it interpretable. By using the attention scores assigned to each word annotation, we can investigate the behavior of the model. Figure \ref{fig:attentions} shows how the attention mechanism focuses on each word in order to estimate the most suitable emoji label.

\section{Conclusion} \label{concl}
% Emojis enhance written language with rich emotional context and can help better decode social media semantic content, thus they start to receive great attention.
In this paper, we present a deep learning system based on a word-level BiLSTM architecture and augment it with contextual attention for SemEval Task 2: \enquote{Multilingual Emoji Prediciction} \cite{semeval2018task2}.
% Furthermore, we use various ensemble schemas to smooth model variations and introduce a novel attention mechanism to highlight the contribution of the most emotionally important words in different contexts.
Our work achieved excellent results, reaching the 2nd place in the competition and outperforming the state-of-the-art reported in the bibliography  \cite{barbieri2017emojis}. The performance of our model could be further boosted, by utilizing transfer learning methods from larger, weakly annotated, datasets. Moreover, the joint training of word- and character-level models can be tested for further performance improvement.  

Finally, we make both our pretrained word embeddings and the source code of our models available to the community\footnote{\url{github.com/cbaziotis/ntua-slp-semeval2018-task2}}, 
in order to make our results easily reproducible and facilitate further experimentation in the field.
\newline

\noindent\textbf{Acknowledgements}. 
This work has been partially supported by the BabyRobot project supported by EU H2020 (grant \#687831). Also, the authors would like to thank NVIDIA for supporting this work by donating a TitanX GPU.

\bibliography{refs}

\begin{thebibliography}{26}
\expandafter\ifx\csname natexlab\endcsname\relax\def\natexlab#1{#1}\fi

\bibitem[{Aoki and Uchida(2011)}]{aoki2011method}
Sho Aoki and Osamu Uchida. 2011.
\newblock A method for automatically generating the emotional vectors of
  emoticons using weblog articles.
\newblock In \emph{Proc. 10th WSEAS Int. Conf. on Applied Computer and Applied
  Computational Science, Stevens Point, Wisconsin, USA}, pages 132--136.

\bibitem[{Bahdanau et~al.(2014)Bahdanau, Cho, and
  Bengio}]{DBLP:journals/corr/BahdanauCB14}
Dzmitry Bahdanau, Kyunghyun Cho, and Yoshua Bengio. 2014.
\newblock \href {http://arxiv.org/abs/1409.0473} {Neural machine translation by
  jointly learning to align and translate}.
\newblock \emph{CoRR}, abs/1409.0473.

\bibitem[{Barbieri et~al.(2017)Barbieri, Ballesteros, and
  Saggion}]{barbieri2017emojis}
Francesco Barbieri, Miguel Ballesteros, and Horacio Saggion. 2017.
\newblock \href {http://aclweb.org/anthology/E17-2017} {Are emojis
  predictable?}
\newblock In \emph{Proceedings of the 15th Conference of the European Chapter
  of the Association for Computational Linguistics: Volume 2, Short Papers},
  pages 105--111. Association for Computational Linguistics.

\bibitem[{Barbieri et~al.(2018)Barbieri, Camacho-Collados, Ronzano,
  Espinosa-Anke, Ballesteros, Basile, Patti, and Saggion}]{semeval2018task2}
Francesco Barbieri, Jose Camacho-Collados, Francesco Ronzano, Luis
  Espinosa-Anke, Miguel Ballesteros, Valerio Basile, Viviana Patti, and Horacio
  Saggion. 2018.
\newblock {SemEval-2018 Task 2: Multilingual Emoji Prediction}.
\newblock In \emph{Proceedings of the 12th International Workshop on Semantic
  Evaluation (SemEval-2018)}, New Orleans, LA, United States. Association for
  Computational Linguistics.

\bibitem[{Barbieri et~al.(2016{\natexlab{a}})Barbieri, Kruszewski, Ronzano, and
  Saggion}]{barbieri2016cosmopolitan}
Francesco Barbieri, German Kruszewski, Francesco Ronzano, and Horacio Saggion.
  2016{\natexlab{a}}.
\newblock How cosmopolitan are emojis?: Exploring emojis usage and meaning over
  different languages with distributional semantics.
\newblock In \emph{Proceedings of the 2016 ACM on Multimedia Conference}, pages
  531--535. ACM.

\bibitem[{Barbieri et~al.(2016{\natexlab{b}})Barbieri, Ronzano, and
  Saggion}]{barbieri2016does}
Francesco Barbieri, Francesco Ronzano, and Horacio Saggion. 2016{\natexlab{b}}.
\newblock What does this emoji mean? a vector space skip-gram model for twitter
  emojis.
\newblock In \emph{LREC}.

\bibitem[{Baziotis et~al.(2017)Baziotis, Pelekis, and
  Doulkeridis}]{baziotis2017datastories}
Christos Baziotis, Nikos Pelekis, and Christos Doulkeridis. 2017.
\newblock Datastories at semeval-2017 task 4: Deep lstm with attention for
  message-level and topic-based sentiment analysis.
\newblock In \emph{Proceedings of the 11th International Workshop on Semantic
  Evaluation (SemEval-2017)}, pages 747--754.

\bibitem[{Bengio et~al.(1994)Bengio, Simard, and Frasconi}]{bengio1994}
Yoshua Bengio, Patrice Simard, and Paolo Frasconi. 1994.
\newblock Learning long-term dependencies with gradient descent is difficult.
\newblock \emph{IEEE transactions on neural networks}, 5(2):157--166.

\bibitem[{Bergstra et~al.(2013)Bergstra, Yamins, and Cox}]{bergstra2013}
James Bergstra, Daniel Yamins, and David~D. Cox. 2013.
\newblock Making a {{Science}} of {{Model Search}}: {{Hyperparameter
  Optimization}} in {{Hundreds}} of {{Dimensions}} for {{Vision
  Architectures}}.
\newblock \emph{ICML (1)}, 28:115--123.

\bibitem[{Cappallo et~al.(2018)Cappallo, Svetlichnaya, Garrigues, Mensink, and
  Snoek}]{cappallo2018new}
Spencer Cappallo, Stacey Svetlichnaya, Pierre Garrigues, Thomas Mensink, and
  Cees~GM Snoek. 2018.
\newblock The new modality: Emoji challenges in prediction, anticipation, and
  retrieval.
\newblock \emph{arXiv preprint arXiv:1801.10253}.

\bibitem[{Cho et~al.(2014)Cho, Van~Merri{\"e}nboer, Gulcehre, Bahdanau,
  Bougares, Schwenk, and Bengio}]{cho2014a}
Kyunghyun Cho, Bart Van~Merri{\"e}nboer, Caglar Gulcehre, Dzmitry Bahdanau,
  Fethi Bougares, Holger Schwenk, and Yoshua Bengio. 2014.
\newblock Learning phrase representations using {{RNN}} encoder-decoder for
  statistical machine translation.
\newblock \emph{arXiv preprint arXiv:1406.1078}.

\bibitem[{Eisner et~al.(2016)Eisner, Rockt{\"a}schel, Augenstein,
  Bo{\v{s}}njak, and Riedel}]{eisner2016emoji2vec}
Ben Eisner, Tim Rockt{\"a}schel, Isabelle Augenstein, Matko Bo{\v{s}}njak, and
  Sebastian Riedel. 2016.
\newblock emoji2vec: Learning emoji representations from their description.
\newblock \emph{arXiv preprint arXiv:1609.08359}.

\bibitem[{Espinosa-Anke et~al.(2016)Espinosa-Anke, Saggion, and
  Barbieri}]{espinosa2016revealing}
Luis Espinosa-Anke, Horacio Saggion, and Francesco Barbieri. 2016.
\newblock Revealing patterns of twitter emoji usage in barcelona and madrid.
\newblock \emph{Frontiers in Artificial Intelligence and Applications.
  2016;(Artificial Intelligence Research and Development) 288: 239-44.}

\bibitem[{Hochreiter et~al.(2001)Hochreiter, Bengio, Frasconi, and
  Schmidhuber}]{hochreiter2001}
Sepp Hochreiter, Yoshua Bengio, Paolo Frasconi, and J{\"u}rgen Schmidhuber.
  2001.
\newblock \emph{Gradient Flow in Recurrent Nets: The Difficulty of Learning
  Long-Term Dependencies}.
\newblock {A field guide to dynamical recurrent neural networks. IEEE Press}.

\bibitem[{Hochreiter and Schmidhuber(1997)}]{hochreiter1997}
Sepp Hochreiter and J{\"u}rgen Schmidhuber. 1997.
\newblock Long short-term memory.
\newblock \emph{Neural computation}, 9(8):1735--1780.

\bibitem[{Jurafsky and James(2000)}]{jurafsky2000}
Daniel Jurafsky and H.~James. 2000.
\newblock Speech and language processing an introduction to natural language
  processing, computational linguistics, and speech.

\bibitem[{Kingma and Ba(2014)}]{kingma2014}
Diederik Kingma and Jimmy Ba. 2014.
\newblock Adam: {{A}} method for stochastic optimization.
\newblock \emph{arXiv preprint arXiv:1412.6980}.

\bibitem[{Ljube{\v{s}}i{\'c} and Fi{\v{s}}er(2016)}]{ljubevsic2016global}
Nikola Ljube{\v{s}}i{\'c} and Darja Fi{\v{s}}er. 2016.
\newblock A global analysis of emoji usage.
\newblock In \emph{Proceedings of the 10th Web as Corpus Workshop}, pages
  82--89.

\bibitem[{Mikolov et~al.(2013)Mikolov, Sutskever, Chen, Corrado, and
  Dean}]{mikolov2013}
Tomas Mikolov, Ilya Sutskever, Kai Chen, Greg~S. Corrado, and Jeff Dean. 2013.
\newblock Distributed representations of words and phrases and their
  compositionality.
\newblock In \emph{Advances in Neural Information Processing Systems}, pages
  3111--3119.

\bibitem[{Pascanu et~al.(2013)Pascanu, Mikolov, and Bengio}]{pascanu2013a}
Razvan Pascanu, Tomas Mikolov, and Yoshua Bengio. 2013.
\newblock On the difficulty of training recurrent neural networks.
\newblock \emph{ICML (3)}, 28:1310--1318.

\bibitem[{Paszke et~al.(2017)Paszke, Gross, Chintala, Chanan, Yang, DeVito,
  Lin, Desmaison, Antiga, and Lerer}]{paszke2017automatic}
Adam Paszke, Sam Gross, Soumith Chintala, Gregory Chanan, Edward Yang, Zachary
  DeVito, Zeming Lin, Alban Desmaison, Luca Antiga, and Adam Lerer. 2017.
\newblock Automatic differentiation in pytorch.

\bibitem[{Pedregosa et~al.(2011)Pedregosa, Varoquaux, Gramfort, Michel,
  Thirion, Grisel, Blondel, Prettenhofer, Weiss, Dubourg, and
  {others}}]{pedregosa2011}
Fabian Pedregosa, Ga{\"e}l Varoquaux, Alexandre Gramfort, Vincent Michel,
  Bertrand Thirion, Olivier Grisel, Mathieu Blondel, Peter Prettenhofer, Ron
  Weiss, Vincent Dubourg, and {others}. 2011.
\newblock Scikit-learn: {{Machine}} learning in {{Python}}.
\newblock \emph{Journal of Machine Learning Research}, 12(Oct):2825--2830.

\bibitem[{{\v R}eh{\r u}{\v r}ek and Sojka(2010)}]{rehurek_lrec}
Radim {\v R}eh{\r u}{\v r}ek and Petr Sojka. 2010.
\newblock {Software Framework for Topic Modelling with Large Corpora}.
\newblock In \emph{{Proceedings of the LREC 2010 Workshop on New Challenges for
  NLP Frameworks}}, pages 45--50, Valletta, Malta. ELRA.
\newblock \url{http://is.muni.cz/publication/884893/en}.

\bibitem[{Saggion et~al.(2017)Saggion, Ballesteros, and
  Barbieri}]{Saggion2017AreEP}
Horacio Saggion, Miguel Ballesteros, and Francesco Barbieri. 2017.
\newblock Are emojis predictable?
\newblock In \emph{EACL}.

\bibitem[{Segaran and Hammerbacher(2009)}]{segaran2009a}
Toby Segaran and Jeff Hammerbacher. 2009.
\newblock \emph{Beautiful {{Data}}: {{The Stories Behind Elegant Data
  Solutions}}}.
\newblock {"O'Reilly Media, Inc."}.

\bibitem[{Srivastava et~al.(2014)Srivastava, Hinton, Krizhevsky, Sutskever, and
  Salakhutdinov}]{srivastava2014}
Nitish Srivastava, Geoffrey~E. Hinton, Alex Krizhevsky, Ilya Sutskever, and
  Ruslan Salakhutdinov. 2014.
\newblock Dropout: A simple way to prevent neural networks from overfitting.
\newblock \emph{Journal of Machine Learning Research}, 15(1):1929--1958.

\end{thebibliography}
\bibliographystyle{acl_natbib}

\appendix

\end{document}